\documentclass{article}
\usepackage{spconf,amsmath,graphicx,hyperref}
\usepackage{multirow}
\usepackage{tabulary} % 在文档开头添加
\usepackage{float}
\usepackage{helvet}
\usepackage[numbers]{natbib}
\usepackage{amsmath}      % 数学公式环境
\usepackage{amssymb}      % 额外的数学符号
\usepackage{algorithm}    % 算法环境
\usepackage{algorithmicx}
\usepackage{algcompatible}  % 算法伪代码环境
\usepackage{caption}      % 自定义图表和算法标题
\usepackage{graphicx}     % 插入图形
\usepackage{hyperref}     % 交叉引用与超链接
\usepackage{bm}           % 粗体数学符号（例如向量）
\usepackage{algpseudocode}
\usepackage[utf8]{inputenc} % allow utf-8 input
\usepackage[T1]{fontenc}    % use 8-bit T1 fonts
\usepackage{hyperref}       % hyperlinks
\usepackage{url}            % simple URL typesetting
\usepackage{booktabs}       % professional-quality tables
\usepackage{amsfonts}       % blackboard math symbols
\usepackage{nicefrac}       % compact symbols for 1/2, etc.
\usepackage{microtype}      % microtypography
\usepackage{xcolor}         % colors
\usepackage{amsmath}        % \usepackage[nonatbib]
\usepackage{booktabs}
\usepackage{tabularx}
\title{SpikingMoE: SDprompt-Guided Dynamic Expert Fusion in Spiking Neural Networks}
\name{%
Yukai Yang$^{1\dagger}$ \quad
Chenxi Qin$^{1\dagger}$ \quad
Jungang Li$^{2\dagger}$ \quad
Xin Zhang$^{3}$ \quad
Wenwei Shao$^{1}$ \quad
Liqun Chen$^{1*}$%
\thanks{This work was supported in part by the National Key R\&D Program of China under Grant No. 2022YFF1202903 (Project: Intelligent Bionic Eye—Information Interaction Design and Implementation).}%
\thanks{$^{\dagger}$These authors contributed equally to this work.}%
\thanks{$^{*}$Corresponding author.}%
}

\address{%
$^{1}$Tianjin University, Medical School, Tianjin 300072, China\\
$^{2}$Southern University of Science and Technology, Department of Computer Science and Engineering,\\
Shenzhen, Guangdong 518055, China\\
$^{3}$Chinese Academy of Medical Sciences, Institute of Biomedical Engineering, Tianjin 300192, China
}

\begin{document}
%\ninept
%
\maketitle
\begin{abstract}

Spiking Neural Networks (SNNs) provide an energy-efficient paradigm for visual recognition. 
We present \textbf{SpikingMoE}, which integrates a spike-driven Transformer with a Mixture-of-Experts (MoE) framework for dynamic computation. 
Inspired by the lateral geniculate nucleus (LGN), a spike-driven prompt (SDprompt) enables input-dependent expert routing in a biologically plausible manner. 
By replacing standard MLPs with spike-compatible expert modules and enforcing binary spike communication, SpikingMoE is designed for neuromorphic hardware. 
Experiments on CIFAR-10 and CIFAR-100 achieve 94.09\% and 74.54\% top-1 accuracy, showing that modular expert routing can be incorporated while retaining reasonable performance. 
To our knowledge, \textbf{SpikingMoE} is the first \emph{open-source} SNN framework that integrates MoE into a spike-driven Transformer with LGN-inspired routing. 
Code is available at the \href{https://github.com/yusgitaa/snnmoe}{Project Page}.

\end{abstract}

\begin{keywords}
Spiking Neural Networks, Mixture-of-Experts, Soft Prompt, Brain-inspired Computing
\end{keywords}
\section{Introduction}
\label{sec:intro}
SNNs, regarded as the third generation of neural networks~\citep{maass1997networks}, emulate the brain's event-driven communication, offering exceptional energy efficiency and biological plausibility on neuromorphic hardware such as Loihi and TrueNorth~\citep{davies2018loihi, akopyan2015truenorth}. By transmitting binary spike signals, SNNs replace energy-intensive multiply-accumulate (MAC) operations with low-power accumulation (AC), achieving up to 100$\times$ energy savings compared to artificial neural networks (ANNs)~\citep{roy2019towards}. Inspired by ANN architectures, SNNs have leveraged structures like ResNet, recurrent networks, and graph neural networks to enhance performance. Recently, the Transformer's self-attention mechanism, excelling in visual tasks such as image classification and object detection, has become a focal point for SNN advancements. Spikformer introduced self-attention to SNNs with Spiking Self-Attention (SSA), using spike-form Query, Key, and Value to avoid softmax and multiplication, enabling efficient sparse computation~\citep{zhou2022spikformer}. The Spike-driven Transformer further proposed Spike-Driven Self-Attention (SDSA), replacing matrix multiplication with Hadamard product and sparse additions, ensuring fully spike-driven computation with significantly reduced energy consumption~\citep{yao2023spike}. However, SNN-Transformer architectures face challenges in scaling to complex visual tasks due to limited model capacity and static computational pathways.

\begin{figure}[htbp]
  \centering
  
  \vspace{-5mm}
  \includegraphics[width=0.5\textwidth]{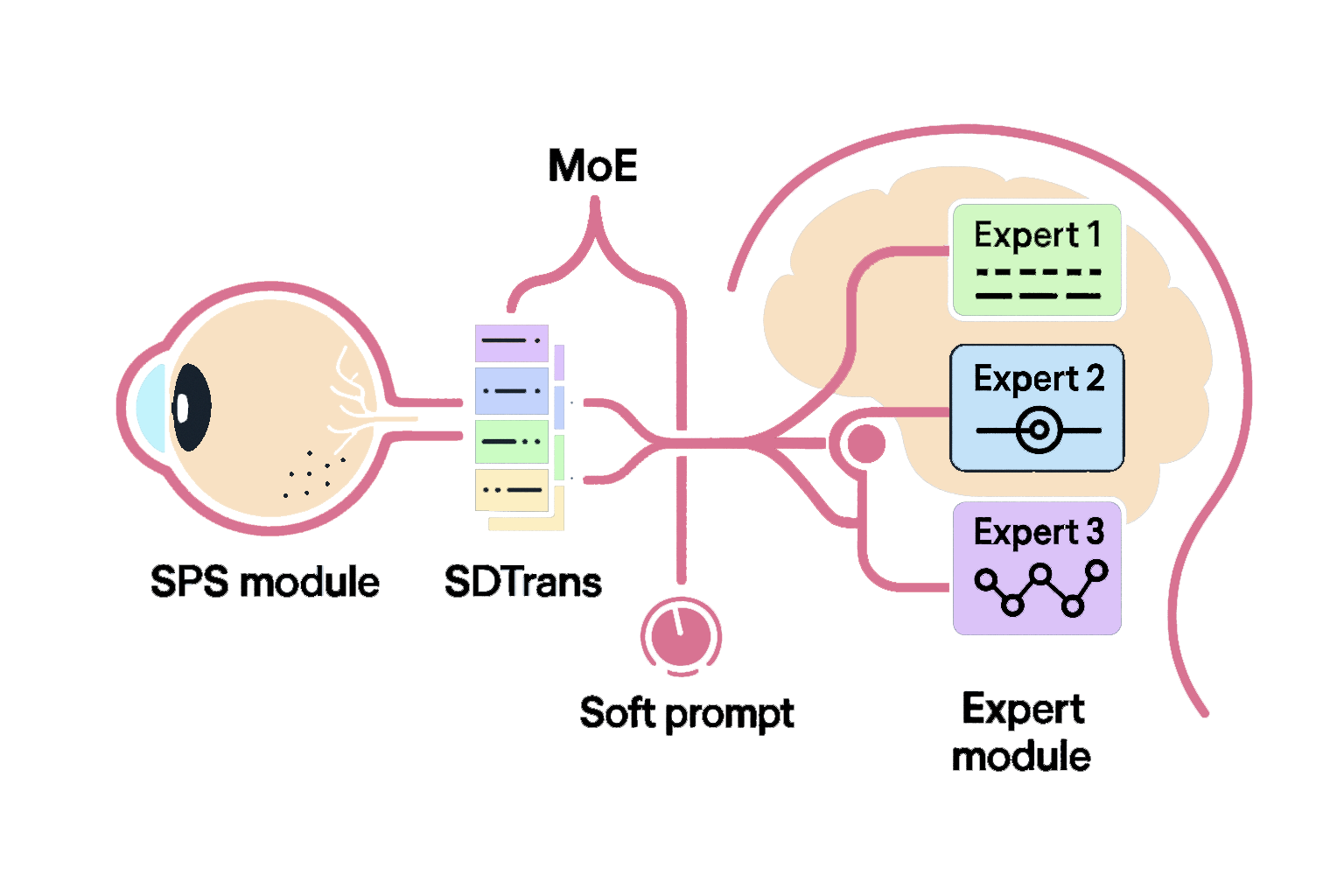}
  \vspace{-5mm}
  \caption{An overview of the spike-driven MoE architecture.}
  \label{fig:overview}
 
\end{figure}

\begin{figure*}[htbp]
  \centering
  
  \includegraphics[width=1\textwidth]{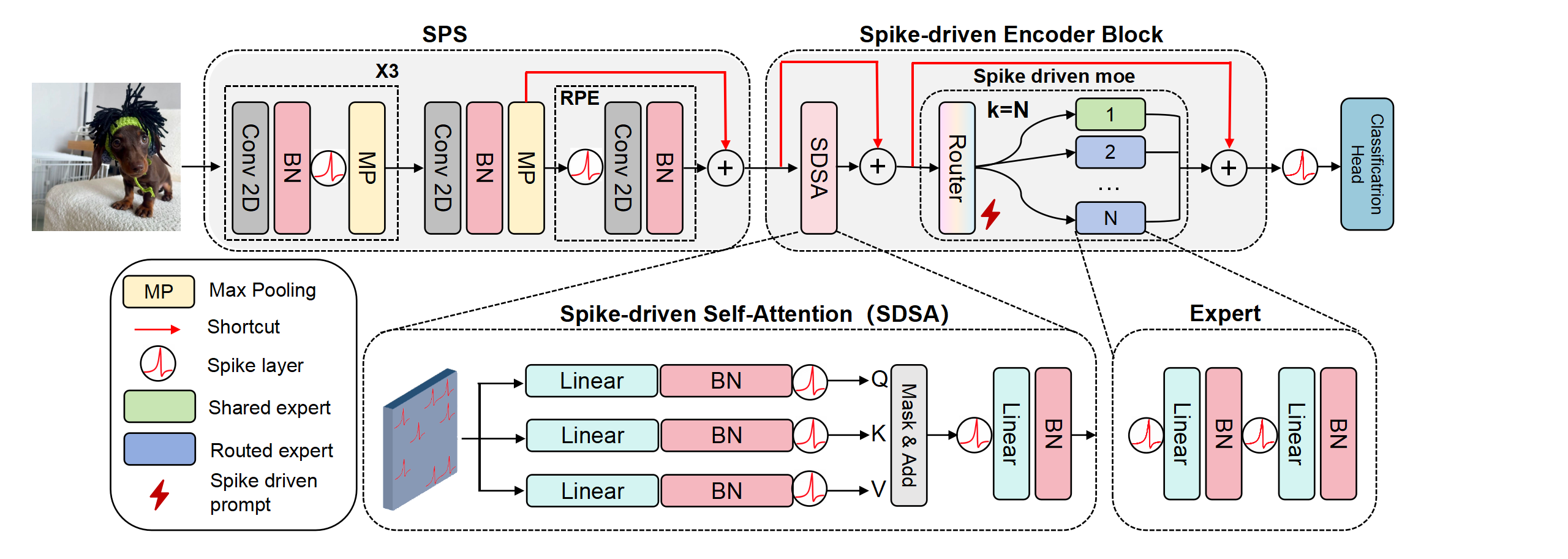}
  \caption{The overview of dynamic expert fusion in SNNs.}
  \label{fig:overview}
   
\end{figure*}
MoE architectures have revolutionized deep learning through sparse expert activation, with models like DeepSeekV3 and LLaMA4 reducing computational costs by up to 50\% while excelling in vision and language tasks~\citep{liu2024deepseek, meta2025llama}. MoE's selective activation mirrors the brain's functional specialization, such as the LGN routing sensory inputs to specific neural pathways, aligning with SNNs' sparse firing patterns. Integrating MoE with SNN-Transformers requires overcoming challenges in spike compatibility and dynamic routing design. To address this, we propose \textbf{SpikingMoE}, a novel architecture that establishes a new paradigm by embedding a spike-compatible MoE framework into the Spike-driven Transformer. 
Inspired by LGN's selective signal modulation, we introduce an SDprompt mechanism for context-aware dynamic expert routing. We replace Spike-driven Transformer's MLP with a spike-compatible MoE model, ensuring fully binary spike communication and sparse addition operations for neuromorphic hardware compatibility. Evaluations on CIFAR-10 and CIFAR-100 show SpikingMoE achieves 94.09\% and 74.54\% top-1 accuracy. SpikingMoE is the first open-source SNN framework to integrate MoE with LGN-inspired SDprompt routing, paving the way for efficient, scalable neuromorphic visual computation.
The contributions of this paper are as follows.
\begin{itemize}
    \item We propose \textbf{SpikingMoE}, a novel architectural paradigm that integrates MoE into SNNs, enabling scalable and modular spiking computation suitable for open-world visual recognition tasks.
    \item We introduce a biologically inspired \textbf{SDprompt-based expert routing} mechanism, motivated by the thalamocortical visual pathway, which facilitates context-dependent and flexible expert selection through a top-down modulatory signal.
    \item Through comprehensive evaluations on both static image and neuromorphic event-based datasets, we demonstrate that SpikingMoE achieves competitive performance while preserving the efficiency and temporal dynamics of SNNs, paving the way for future research in dynamic, modular spiking architectures.
\end{itemize}

\section{Related Work}
\label{sec:Related Work}

Spiking Neural Networks (SNNs) offer a biologically inspired, energy-efficient alternative via event-driven computation. Recent SNN–Transformer variants—Spikformer~\citep{zhou2022spikformer} and the Spike-driven Transformer~\citep{yao2023spike}—adapt self-attention to spikes, replacing softmax/multiplications with spike-domain operations, thereby reducing energy while retaining competitive accuracy.

Mixture-of-Experts (MoE) scales deep models by sparsely activating specialized experts (e.g., DeepSeekV3~\citep{liu2024deepseek}). This selectivity mirrors neural specialization and aligns with SNN sparsity, yet MoE within spiking frameworks remains underexplored due to spike-compatible routing and expert design.

Prompting further enables adaptive computation: both manual prompts~\citep{guo2023calip,li2022language,radford2021learning,xu2022groupvit,yong2023prompt} and learnable variants~\citep{guo2023texts,rao2022denseclip,zhou2022conditional,zhou2022learning} are effective, but most rely on dense floating-point operations, limiting applicability to SNNs.

We introduce \textbf{SpikingMoE}, the framework that combines spike-driven Transformers with MoE via a biologically inspired spike-driven prompt (SDprompt), enabling input-dependent, context-aware expert routing with fully spike-compatible operations.

\vspace{-2mm}
\section{Method}

\label{sec:Method}
We present \textbf{SpikingMoE}, an extension of the spike-driven Transformer that integrates a spike-compatible Mixture-of-Experts (MoE) with an SDprompt mechanism for dynamic routing. The design preserves event-driven efficiency while enabling input-dependent specialization.

\subsection{Overall Architecture}
Given an input sequence \(I\), the Spiking Patch Splitting (SPS) module produces spike-form patch embeddings \(S_0\). The encoder stacks \(L\) layers, each composed of Spike-Driven Self-Attention (SDSA) and a spike-compatible MoE block with SDprompt-based gating. Residual/membrane shortcuts maintain binary spike communication throughout. Global average pooling followed by a linear classifier yields the final prediction.

\subsection{Spiking Neuron Model}
We adopt the Leaky Integrate-and-Fire (LIF) neuron:
\begin{align}
U[t] &= H[t-1] + X[t], \\
S[t] &= \operatorname{Hea}(U[t]-u_{\text{th}}), \\
H[t] &= V_{\text{reset}} S[t] + (\beta U[t]) (1-S[t]),
\end{align}
where \(U[t]\) is the membrane potential, \(H[t]\) the state, \(S[t]\in\{0,1\}\) the spike, and \(\beta<1\) the decay factor. 
This add-and-threshold formulation keeps computation fully spike-driven (additions and masks) without multiply–accumulate operations.

\vspace{-2mm}
\subsection{Spike-Driven Self-Attention (SDSA)}
\label{sec:sdsacomp}
Given \(S_{l-1}\in\{0,1\}^{T\times N\times D}\), spike-linear projections produce \(Q_S,K_S,V_S\).
Attention is computed by spike-domain correlations and channel summation:
\begin{equation}
\operatorname{SDSA}(Q,K,V)=\operatorname{SN}\!\big(\operatorname{SUM}_c(Q_S\otimes K_S)\big)\otimes V_S,
\end{equation}
where \(\otimes\) is the Hadamard product and \(\operatorname{SUM}_c\) sums over channels.
This yields linear-time, addition-dominant operations compatible with neuromorphic execution.

\vspace{-2mm}
\subsection{MoE Layer with SDprompt Routing}
We replace the MLP in each encoder layer with a spike-compatible MoE block comprising \(K\) experts and an SDprompt-enhanced gate (Sec.~\ref{sec:SDprompt}). 
For each token, the gate selects the top-\(k\) experts, which are evaluated independently; thus only \(k\) of \(K\) experts are active per token (sparse computation).
\textbf{Each expert is a two-layer spike MLP.}
The MoE output for token \(n\) is
\begin{equation}
U_{\text{MoE},n}=\frac{1}{k}\sum_{m=1}^k \mathrm{Expert}_{i_m}(S'_{l,n}),\quad i_m\in \mathrm{selected}_n,
\end{equation}
followed by residual merging and spike normalization.

\subsubsection{Auxiliary Routing Loss}
To prevent expert collapse and promote balanced, diverse routing, we add
\begin{equation}
\mathcal{L}_{\text{aux}}=\alpha_{\text{aux}}\big(\mathcal{L}_{\text{balance}}-\mathcal{L}_{\text{importance}}\big).
\end{equation}
Here the load-balancing term is 
$\mathcal{L}_{\text{balance}} = n_{\text{routed}}\cdot \mathrm{MSE}(\mathbf{u},\mathbf{u}^*)$, 
where $u_k=\frac{c_k}{\sum_j c_j + 1\mathrm{e}{-}8}$ with $c_k$ the routed-token count for expert $k$, and $\mathbf{u}^*=(1/n_{\text{routed}})\mathbf{1}$ over active experts.

\subsubsection{SDprompt-Enhanced Routing}\label{sec:SDprompt}
SDprompt conditions the gate on input context. A spike-linear module fuses \(S'_l\) with prompt \(P\) and emits binary gating spikes:
\begin{equation}
G=\operatorname{SN}\!\big(\operatorname{Linear}_{\mathrm{spike}}([S'_l;P])\big)\;,\quad G\in\{0,1\}^{T\times N\times K}.
\end{equation}
For each token, the top-\(k\) experts are chosen by accumulated gating activity over time, 
\(\mathrm{selected}_n=\operatorname{top}\text{-}k\!\big(\sum_{t=1}^{T}G_{t,n,:}\big)\).

\section{Experiments}
\begin{table}[t]
\centering
\caption{Top‐1 classification accuracy (\%) of our model on CIFAR-10 and CIFAR-100 datasets. (Train 400 epochs)}
\label{tab:table1}
\footnotesize 
\setlength{\tabcolsep}{2.5pt} 
\begin{tabular}{@{}l l c c c c@{}} 
\toprule
% --- 通过修改 \\ 为 \\[-2pt] 来缩短嵌套单元格内的行距 ---
\begin{tabular}[c]{@{}l@{}}\textbf{Method}\end{tabular} & 
\begin{tabular}[c]{@{}l@{}}\textbf{Architecture}\end{tabular} & 
\begin{tabular}[c]{@{}c@{}}\textbf{Time}\\[-3pt]\textbf{steps}\end{tabular} &  
\begin{tabular}[c]{@{}c@{}}\textbf{CIFAR-10}\\[-3pt]\textbf{Acc}\end{tabular} & 
\begin{tabular}[c]{@{}c@{}}\textbf{CIFAR-100}\\[-2pt]\textbf{Acc}\end{tabular} \\
\midrule
Hybrid training & VGG-11 & 125 & 92.22 & 67.87 \\
Diet-SNN & ResNet-20 & 10/5  & 92.54 & 64.07 \\
STBP & CIFARNet & 12  & 89.93 & - \\
STBP NeuNorm & CIFARNet & 12 & 90.53 & - \\
TSSL-BP & CIFARNet & 4 & 91.41 & - \\
STBP-tdBN & ResNet-19 & 4 & 92.92 & 70.86 \\
TET & ResNet-19 & 4 & \textbf{94.44} & 74.47 \\
MS-ResNet & ResNet-110 & - & 91.72 & 66.83 \\
& ResNet-482 & - & 91.90 & - \\
PLIF & SNN & 8 & 93.50 & - \\
Dspike & ResNet18 & 6 & 94.30 & 74.54 \\
\midrule
% --- 这里同样缩短行距 ---
\begin{tabular}[c]{@{}l@{}}Spike-Driven\\[-1pt]Transformer\end{tabular} & 2-256 & 4 & \underline{94.38} & \textbf{75.61} \\
Ours & 2-256 & 4 & 92.65 & 72.06 \\
Ours & 2-256 & 8 & 93.58 & \underline{75.00} \\
Ours & 2-384 & 4 & 94.02 & 74.45 \\
Ours & 4-256 & 4 & 92.53 & 72.03 \\
Ours & 4-384 & 4 & 94.09 & 74.54 \\
\bottomrule
\end{tabular}
\end{table}

\begin{table}[t]
\centering
\caption{Top‐1 classification accuracy (\%) of our model on CIFAR-10 DVS and Gesture-DVS datasets.}
\label{tab:table2}
\footnotesize % 1. 使用小号字体
\setlength{\tabcolsep}{3.5pt} % 2. 调整列间距 (可根据需要微调)
\begin{tabular}{@{} l l cc cc @{}} % 3. 移除表格外侧空白
\toprule
% --- 4. 重构并简化表头 ---
\multirow{2}{*}{\textbf{Method}} & \multirow{2}{*}{\textbf{Architecture}} & \multicolumn{2}{c}{\textbf{CIFAR-10 DVS}} & \multicolumn{2}{c}{\textbf{Gesture-DVS}} \\
\cmidrule(lr){3-4} \cmidrule(lr){5-6}
& & \textbf{Acc} & \textbf{Steps} & \textbf{Acc} & \textbf{Steps} \\
\midrule
LIAF-Net & VGG & 70.4 & 10 & 97.6 & 60 \\
TA-SNN & CNN & 72.0 & 10 & \textbf{98.6} & 60 \\
Rollout & DenseNet & 66.8 & 48 & 97.2 & 240 \\
DECOLLE & CNN & - & - & 95.5 & 500 \\
STBP-tdBN & ResNet-19 & 67.8 & 10 & 96.9 & 40 \\
PLIF & CNN & 74.8 & 20 & 97.6 & 20 \\
SEW-ResNet & ResNet-18 & 74.4 & 16 & 97.9 & 16 \\
Dspike & ResNet-18 & 75.4 & 10 & - & - \\
DSR & VGG-11 & 77.3 & 10 & - & - \\
Slayer & CNN & - & - & 93.6 & - \\
\midrule
% --- 5. 对长方法名进行换行和行距压缩处理 ---
\begin{tabular}[c]{@{}l@{}}Spike-Driven\\[-1pt]Transformer\end{tabular} & 2-256 & 74.00 & 10 & 95.83 & 10 \\
\begin{tabular}[c]{@{}l@{}}Spike-Driven\\[-1pt]Transformer\end{tabular} & 2-512 & 73.70 & 10 & 97.56 & 10\\
\begin{tabular}[c]{@{}l@{}}Spike-Driven\\[-1pt]Transformer\end{tabular} & 2-512 & \underline{76.30} & 16 & \textbf{98.61} & 16\\
Ours & 2-256 & 66.30 & 10 & 94.44 & 10 \\
Ours & 2-256 & 72.70 & 16 & 95.83 & 16 \\
Ours & 2-512 & 67.20 & 10 & 92.70 & 10 \\
Ours & 2-512 & \textbf{77.49} & 16 & \underline{97.91} & 16 \\
\bottomrule
\end{tabular}
\end{table}

\begin{table}[t]
  \centering
  \caption{Ablation of MoE and SDprompt (top-1\%). Baseline: spike-driven Transformer.}
  \label{tab:table3}
  \footnotesize                               % 比 normal 小一号，通常即可避免溢出
  \setlength{\tabcolsep}{2.8pt}               % 缩小列间距
  \renewcommand{\arraystretch}{1.05}          % 轻微增大行距，保证可读性
  \begin{tabular}{@{}lrrrr@{}}                % @{} 去掉左右额外空白；r 让数字更紧凑
    \toprule
    \textbf{Model} & \textbf{Cifar-10} & \textbf{Cifar-100} & \textbf{Cifar-10 DVS} & \textbf{Gesture-DVS} \\
    \midrule
    Baseline            & 94.38 & 75.61 & 76.30 & 98.61 \\
    +MoE                & 92.07 {\scriptsize(↓2.31)} & 73.42 {\scriptsize(↓2.19)} & 73.10 {\scriptsize(↓3.20)} & 97.20 {\scriptsize(↓1.41)} \\
    +MoE+SDprompt       & 92.65 {\scriptsize(↑0.58)} & 73.61 {\scriptsize(↑0.19)} & 77.49 {\scriptsize(↑4.39)} & 97.91 {\scriptsize(↑0.71)} \\
    \bottomrule
  \end{tabular}
\end{table}

% \begin{table*}[t]
%   \centering
%   \caption{Top-1 classification accuracy (\%) of our model on CIFAR-10 and CIFAR-100 datasets with different modules.}
%   \label{tab:table3}
%   \begin{tabulary}{\textwidth}{LCCCJ} % L, C, J 是 tabulary 的列类型
%     \toprule
%     Model & \shortstack{CIFAR-10 \\ Acc}& \shortstack{CIFAR-100 \\ Acc} & \shortstack{CIFAR-10 DVS \\ Acc}& \shortstack{Gesture DVS \\ Acc} \\
%     \midrule
%     Baseline &  94.38 & 75.61 &  76.30 & 98.61 \\
%     Baseline+MoE &  92.07 & 73.42 & 64.7 &92.70\\
%     Baseline+MoE+SDPrompt & 92.65 & 73.61& 77.49 &97.91 \\
%     \bottomrule
%   \end{tabulary}
% \end{table*}
\label{sec:experiments}

We evaluate SpikingMoE on four benchmarks: CIFAR-10, CIFAR-100, CIFAR10-DVS, and DVS128 Gesture, covering both static image classification and neuromorphic event-based recognition.

\textbf{Training configuration.}
For CIFAR-10/100 and CIFAR10-DVS we use AdamW, while LAMB is adopted for Gesture for stability; all models are trained with cosine schedules and warm-up, label smoothing (0.1) and weight decay with dataset-specific learning rates, and run for 300 epochs on CIFAR and 200 on DVS with AMP enabled on DVS.

\textbf{Data augmentation and regularization.}
CIFAR uses bicubic resize, horizontal flip, RandAugment, Mixup (disabled in late training), and color jitter, while DVS datasets use only resize and flip without color-based or AutoAugment strategies, following common SNN practice.

\textbf{Model configuration.}
All models adopt a spike-driven Transformer with 2 encoder layers (256-dim, 8 heads); MLPs are replaced by MoE modules with four experts, with top-2 selected per token via SDprompt-based routing. Inputs are \(32\times32\) (CIFAR-10/100), \(64\times64\) (CIFAR10-DVS), and \(128\times128\) (Gesture), with temporal lengths 4 (CIFAR) and 10 (DVS); CIFAR uses 3 channels and DVS uses 2-channel events. For DVS, linear layers in attention/experts are swapped with convolutional substitutes, and Temporal Efficient Training (TET) is used only for CIFAR10-DVS; other hyperparameters (e.g., dropout, drop-path) follow defaults.

\textbf{MoE configuration.}
Each MoE layer has \(K=4\) experts (three unique plus one shared across layers); at each timestep an SDprompt-enhanced gate selects top-\(k=2\) experts. To regularize usage we apply an auxiliary routing loss combining load balancing and an entropy-based importance term with weight \(\alpha_{\text{aux}}=0.1\).

\begin{figure}[htbp]
  \centering
  \includegraphics[width=0.48\textwidth]{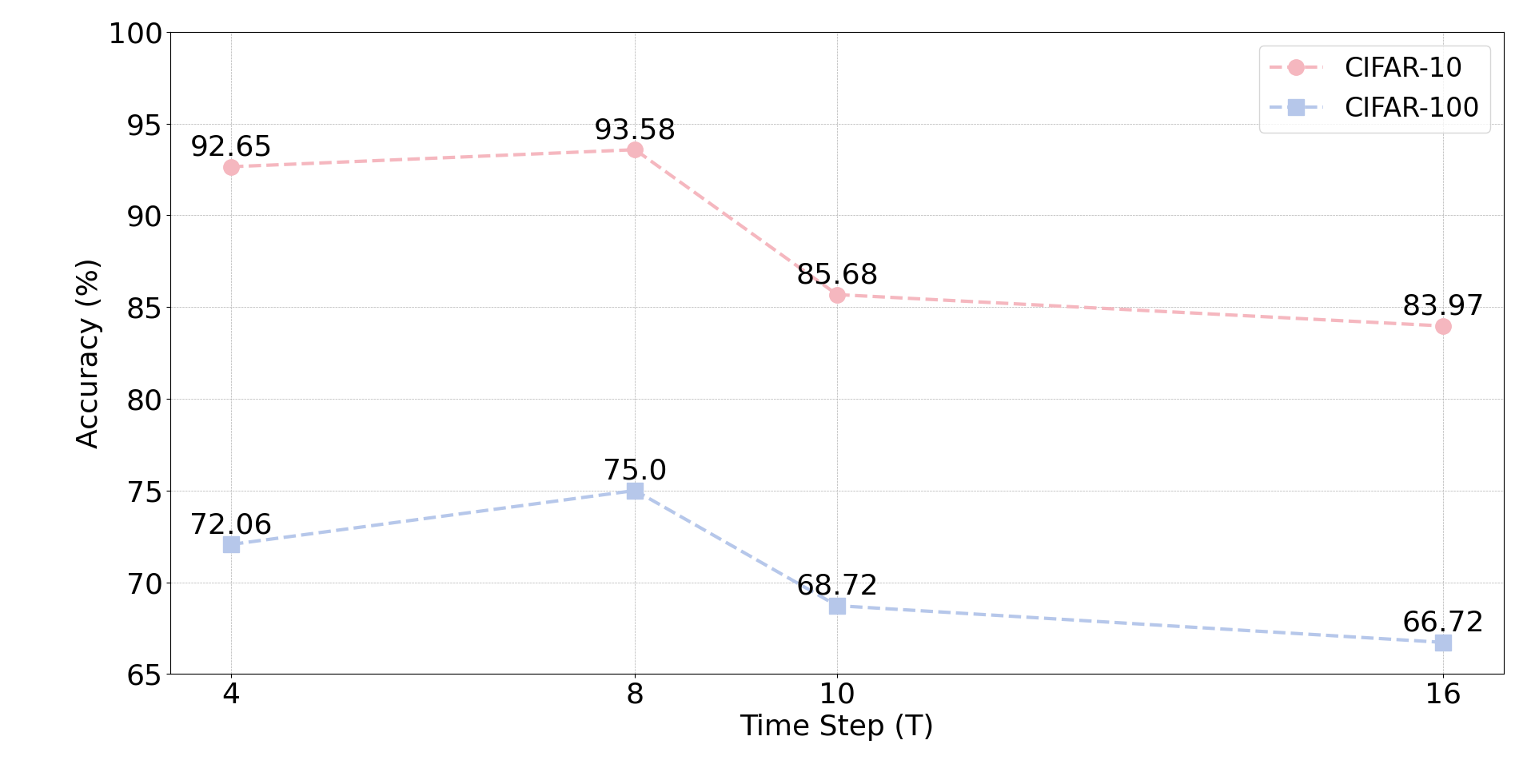}
  \caption{Performance of SpikingMoE Across Different Time Steps.}
  \label{fig:timestep}
  \vspace{-3mm}
\end{figure}

\subsection{Comparison with Prior Methods}
Tables~\ref{tab:table1} and \ref{tab:table2} summarize the results of SpikingMoE compared with representative spiking and non-spiking models. On CIFAR-10 and CIFAR-100, our model reaches 94.09\% and 74.54\% top-1 accuracy, which is comparable to prior ResNet-based SNNs such as Dspike~\citep{li2021differentiable} and TET~\citep{deng2022temporal}. On CIFAR10-DVS and Gesture, it achieves 72.70\% and 95.83\%, showing that dynamic expert routing can be incorporated into spike-driven Transformers while retaining competitive performance. These results suggest that integrating MoE with SNNs is feasible, and provide an initial step toward modular and adaptive spike-based architectures.

\subsection{Ablation Study}
We assess component contributions in Table~\ref{tab:table3}. Using MoE alone moderately reduces accuracy, consistent with added routing complexity; adding SDprompt recovers part of the drop and improves Gesture, suggesting that context-dependent prompts stabilize expert selection. Empirical results in Fig.~\ref{fig:timestep} further show a distinct non-monotonic dependence on the number of time steps \(T\): accuracy improves at moderate \(T\) but degrades for large \(T\) due to spike saturation and noisier routing. Attention maps (Fig.~\ref{fig:overview}) indicate focus on salient object regions, supporting interpretability under spiking constraints.

\begin{figure}[htbp]
  \centering
  \includegraphics[width=0.45\textwidth]{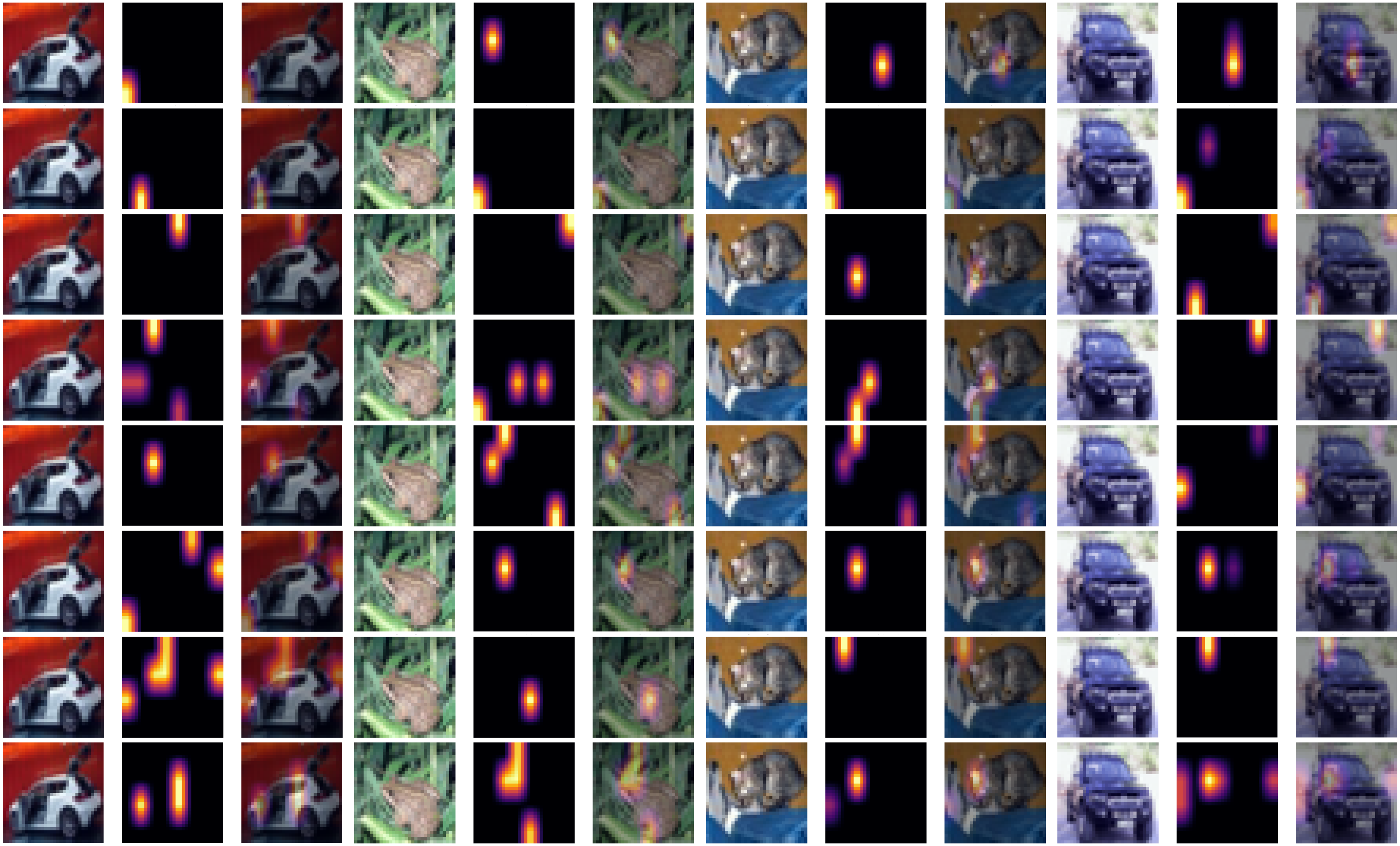}
  \caption{Attention maps from different heads in the SSA.}
  \label{fig:overview}
  \vspace{-3mm}
\end{figure}

% \begin{figure}[htbp]
%   \centering
%   \includegraphics[width=0.45\textwidth]{figures/3.png}
%   \caption{Attention maps from different heads in the SSA.}
%   \label{fig:overview}
% \end{figure}

\vspace{-2mm}
\section{Conclusion}
\label{sec:Conclusion}
We presented \textbf{SpikingMoE}, integrating a spike-compatible Mixture-of-Experts into a spike-driven Transformer via an \textbf{SDprompt} for context-dependent routing. Although gains are not uniform across benchmarks, our results show MoE can be incorporated into spiking models to enable modular, dynamic computation. This work is an initial step toward biologically inspired, scalable, and interpretable neuromorphic vision, motivating future studies on energy, scaling, and broader datasets.

% \section{Acknowledgment}
% \label{sec:Acknowledgment}
% This study was supported by the National Key
% Research and Development Program of China [grant numbers: 2022YFF1202900],
% the National Natural Science Foundation of China [grant numbers: 82471196 ]

\section{References}
\label{sec:References}
\begingroup
\renewcommand{\section}[2]{}% 只在组内屏蔽 BibTeX 的自动标题
\bibliographystyle{IEEEbib}   % 或 IEEEtran（按会议要求）
\bibliography{strings,refs}
\endgroup

\end{document}